\documentclass[sigconf]{acmart}

\AtBeginDocument{%
  }

\copyrightyear{2024} 
\acmYear{2024} 
\setcopyright{rightsretained} 
\acmConference[CSCW Companion '24]{Companion of the 2024 Computer-Supported Cooperative Work and Social Computing}{November 9--13, 2024}{San Jose, Costa Rica}
\acmBooktitle{Companion of the 2024 Computer-Supported Cooperative Work and Social Computing (CSCW Companion '24), November 9--13, 2024, San Jose, Costa Rica}\acmDOI{10.1145/3678884.3681831}
\acmISBN{979-8-4007-1114-5/24/11}

\usepackage{enumitem}
\usepackage{colortbl}

\begin{document}

\title{From Stem to Stern: Contestability Along AI Value Chains}
\author{Agathe Balayn}
\email{a.m.a.balayn@tudelft.nl}
\affiliation{%
  \institution{Delft University of Technology}
  \city{Delft}
  \country{The Netherlands}
}
\authornote{Both authors contributed equally to the paper}

\author{Yulu Pi}
\email{yulu.pi@warwick.ac.uk}
\affiliation{%
  \institution{University of Warwick}
  \city{Coventry}
  \country{United Kingdom}
}\authornotemark[1]

\author{David Gray Widder}
\email{david.g.widder@gmail.com}
\affiliation{%
  \institution{Cornell University}
  \country{United States of America}
}
\author{Kars Alfrink}
\email{c.p.alfrink@tudelft.nl}
\affiliation{%
  \institution{Delft University of Technology}
  \city{Delft}
  \country{The Netherlands}
}

\author{Mireia Yurrita}
\email{m.yurritasemperena@tudelft.nl}
\affiliation{%
  \institution{Delft University of Technology}
  \city{Delft}
  \country{The Netherlands}
}

\author{Sohini Upadhyay}
\email{supadhyay@g.harvard.edu}
\affiliation{%
  \institution{Harvard University}
  \city{}
  \country{United States of America}
}

\author{Naveena Karusala}
\email{naveenak@seas.harvard.edu}
\affiliation{%
  \institution{Harvard University}
  \city{}
  \country{United States of America}
}

\author{Henrietta Lyons}
\email{hlyons@student.unimelb.edu.au}
\affiliation{%
  \institution{University of Melbourne}
  \city{Melbourne}
  \country{Australia}
}

\author{Cagatay Turkay}
\email{Cagatay.Turkay@warwick.ac.uk}
\affiliation{%
  \institution{University of Warwick}
  \city{}
  \country{United Kingdom}
}
\author{Christelle Tessono}
\email{christelle.tessono@mail.utoronto.ca}
\affiliation{%
  \institution{University of Toronto}
  \city{}
  \country{Canada}
}

\author{Blair Attard-Frost}
\email{blair@blairaf.com}
\affiliation{%
  \institution{University of Toronto}
  \city{}
  \country{Canada}
}

\author{Ujwal Gadiraju}
\email{u.k.gadiraju@tudelft.nl}
\affiliation{%
  \institution{Delft University of Technology}
  \city{Delft}
  \country{The Netherlands}
}
\renewcommand{\shortauthors}{Balayn, Pi, et al.}

\begin{abstract} 
This workshop will grow and consolidate a community of interdisciplinary CSCW researchers focusing on the topic of contestable AI. As an outcome of the workshop, we will synthesize the most pressing opportunities and challenges for contestability along AI value chains in the form of a research roadmap. This roadmap will help shape and inspire imminent work in this field. Considering the length and depth of AI value chains, it will especially spur discussions around the contestability of AI systems along various sites of such chains. The workshop will serve as a platform for dialogue and demonstrations of concrete, successful, and unsuccessful examples of AI systems that (could or should) have been contested, to identify requirements, obstacles, and opportunities for designing and deploying contestable AI in various contexts.
This will be held primarily as an in-person workshop, with some hybrid accommodation. The day will consist of individual presentations and group activities to stimulate ideation and inspire broad reflections on the field of contestable AI. Our aim is to facilitate interdisciplinary dialogue by bringing together researchers, practitioners, and stakeholders to foster the design and deployment of contestable AI.

\end{abstract}





\maketitle
\section{Background}

In recent years, the Computer-Supported Cooperative Work (CSCW), Human-Computer Interaction (HCI), and Artificial Intelligence (AI) communities have become interested in contestable AI as a means to confront, acknowledge, and rectify the negative impacts caused by AI systems. Contestable AI refers to AI systems that are open and responsive to human dispute and intervention throughout their lifecycle \cite{alfrink_contestable_2022}. This interest is evident in theoretical and empirical research and practice \cite{Lyons_2021,Ploug2020TheFD,almada_human_2019,Kluttz_Kohli_Mulligan_2020}, as well as AI governance initiatives that aim to explore contestability as a means to enhance human agency \cite{fanni_enhancing_2023} and address ethical and societal implications of AI. For example, consider the 2020 United Kingdom school exam grading controversy \cite{grading-fiasco}: students protested the use of an AI algorithm to determine their grades, holding signs that read ``Your algorithm does not know me.'' This highlighted the urgent need for AI systems and processes that are open to human intervention and responsive to disputes.

Contestable AI is a growing interdisciplinary field. Legal scholars have proposed the right to contest \cite{bayamlioglu_right_2022}, which ensures a level of protection for individuals affected by algorithmic decisions \cite{kaminski2021right}. Meanwhile, HCI and CSCW researchers view contestability from a design perspective, focusing on making AI systems contestable to developers and end users by design \cite{alfrink_contestable_2022,alfrink_contestable_2023,karusala2024understanding,yurrita2023generating}. While these efforts have made significant progress in directing the conversation towards making AI more responsive and accountable 
through ongoing learning 
based on feedback and contestation \cite{mulligan2019procurement}, their main focus remains on contesting AI design or outputs \cite{AttardFrost_2023} within specific domains such as content moderation \cite{content}. 

A broader perspective on contestability can be gained by considering AI systems as dynamic sociotechnical systems with temporal and spatial dimensions. This approach expands our horizons to consider contestable AI as a \textit{value chain problem}\footnote{The literature uses the terms ``AI value chain'', e.g., \cite{madiega2021artificial}, and ``AI supply chain'', e.g., \cite{widder2023dislocated,cobbe2023understanding}. We use them interchangeably in this workshop.} 
\cite{cobbe2023understanding,cobbe2024governance,widder2023dislocated}). This encompasses the entirety of the AI lifecycle, including various actions taken by different actors: the extraction of materials, the construction of physical infrastructures, the decision-making process in data collection, model development, modes of human oversight and the impact on individuals, society and the environment. CSCW methodologies and prior insights around collaborative work are of particular relevance to investigating the AI value chain through the lens of contestability.

Exploring contestable AI along the AI value chain broadens the scope of potential sites and angles for contestation. For instance, harms arising further up the AI value chain, such as poor labor conditions, negative environmental impacts, 
and the inappropriate collection and use of personal data, remain relatively absent from the academic discourse around contestable AI. In recent years, activist efforts have emerged in response to the growing concerns surrounding AI systems and their impact on society. Should those activities fall into our discussion of contestable AI, and if so, how? One notable example is the Data Centre Activism in Chile, Ireland, and the Netherlands \cite{lehuede2024elemental}. Activists in these countries have protested against the construction and expansion of data centers, citing concerns about their environmental impact, energy consumption, and the potential for AI systems to exacerbate social inequalities. These real-life examples show that contestability could entail public discourse and policy debates, involving individuals utilizing social media platforms to raise their concerns or actively engaging in decisions around AI. However, there is a lack of discussion regarding how these practices fit within the scope of contestable AI and the unique challenges and opportunities faced. 

Moreover, adopting the value-chain perspective in contestable AI exposes and prompts discussion of the inherent challenges of attributing responsibility. Contesting AI may face the ``many hands problem,'' as the design or output being contested may result from a chain of different actors contributing in different ways and capacities to the production, deployment, and use 
\cite{cobbe2023understanding}. This raises crucial questions: what and who exactly are we contesting, and who precisely bears the responsibility to address the contestation? It also opens the door to moving beyond individualistic approaches, and exploring how communities or society can contest AI collectively. 

To address these issues and support this emerging research area, the goal of this workshop is to explore the impacts, challenges, opportunities, and limitations of contestable AI along the AI value chain around the world. We are particularly interested in empirical studies, including real-world case studies and user studies with both positive and negative outcomes, along with reflections on lessons learned. Additionally, we seek descriptions of unique contexts for contestable AI and their varied and specific challenges, and visionary discussions, and proposals about the implications of contestable AI on the design and governance of AI systems. Focusing on real-world scenarios and the entire AI value chain provides a unique, concrete, and more comprehensive perspective on contestable AI, different from previous workshops on contestability \cite{contestability}. 

\section{Workshop Goal}
Through a series of talks, keynotes, and group work, we expect the workshop to achieve the following outcomes:

\begin{itemize}[leftmargin=*]
  \item Developing a holistic understanding of the requirements, challenges, existing support, and forthcoming opportunities to inform future research and practice on contestable AI.
  \item Identifying and synthesizing unique contributions of the CSCW community in contestable AI.
    \item Fostering an interdisciplinary community of researchers and practitioners spanning  
    relevant disciplines (e.g., HCI, AI, law, economics, political science, among others).
    \item Developing shared vocabulary and priorities across disciplines and communities to design for contestability along AI chains.
    \item 
    Publishing an article in the Communications of the ACM (CACM) sketching out a roadmap for contestable AI research that can inspire and shape research in this nexus over the next 10 years.
\end{itemize}

\section{Workshop structure}

\subsection{Workshop schedule}

We schedule the workshop for one day, as outlined in Table \ref{tab:schedule}. 
After the workshop, the organizers and interested participants will gather (virtually),  
 generate a summary of the key insights gleaned from the workshop discussions, and craft a roadmap for future research.

\begin{table}[h]
    \centering   
    \caption{Tentative Workshop Schedule}
    \label{tab:schedule}
    \begin{tabular}{l p{6.2cm}}
        \toprule
         09:00--09:20 & Introduction of the workshop organizers \& goals \\
        09:20--09:50 & 
       Keynote 1 \\
       09:50--10:35 &  Participant presentations 1 \\
         \cellcolor[gray]{.8}10:35--10:50 & \cellcolor[gray]{.8}Coffee break \\
         10:50--11:40 & Participant presentations 2 \\
         11:40--12:00 & Introduction of the group activities\\
         12:00--12:45 & 
         [GA1] Mapping and brainstorming \\
         \cellcolor[gray]{.8}12:45--14:00& \cellcolor[gray]{.8}Lunch \\
         14:00--14:30& Keynote 2  \\
         14:30--15:30& [GA2] Deep-dive on sites of the AI supply chain\\
         \cellcolor[gray]{.8}15:30--15:45& \cellcolor[gray]{.8}Coffee break \\
         15:45--17:00& [GA3] Deep dive on transversal themes \\
         17:00--17:30 & Synthesis of the workshop's insights \\
     17:30--17:45 & Open discussion and closing \\
        \bottomrule
    \end{tabular}
\end{table}

\subsection{Activity Description}

The workshop will consist of the following activities. 

\textbf{Participant presentation:} 
We will ask each participant to present an overview of their submission. Depending on the number of submissions, we will accommodate the length of the presentations to create room for every submission to be presented. 


\textbf{Keynote:} 
We will invite keynote speakers whose insightful perspectives may not be fully represented in the submissions received. These speakers may include individuals from civil society or legal scholars, with whom we are already in contact.

\textbf{GA1: Mapping of the presentations onto sites of the value chain and brainstorming:} Following participants' presentations, we will ask them to use a physical and / or virtual board to place their work on a map for contestable AI. This map will be preliminarily prepared by the workshop organizers (--who, what, why, when, how--, and --challenges, needs, opportunities-- across the sites of the AI value chain), and will serve to kick off discussions. 
We will encourage participants to conduct the activity collaboratively, discuss potential challenges and disagreements they face, observe similarities between their use cases, and identify additional relevant dimensions and extensions to the map. This exercise will also be useful to identify blind spots in the contestable AI space. 

\textbf{GA2: Deep-dive on contestability sites:} 
We will ask participants to group themselves based on the sites in the map for which they have the most experience, and to start discussing commonalities, differences, and challenges and opportunities they face. 
We will then ask them to report on their discussions, and to make the collective exercise of identifying recurring topics across the sites.

\textbf{GA3: Deep-dive on transversal themes:} The third group exercise will consist of delving deeper into certain recurring (transversal) topics that come out from the presentations and group activities. We will again encourage participants to reflect on the needs and opportunities for each of these themes and ideate on how to research these topics. Examples of such themes are: individual versus collaborative contestability, contestability within versus contestability from the outside of the supply chain, ultimate goals of contestability and relevant means for contesting, the specificities of contestability across geographic and cultural locations.

\textbf{Synthesizing a roadmap for future research:} In plenary sessions and smaller groups, we will accumulate the findings of the workshop, and devise the best way to synthesize them into a research roadmap for contestable AI. 
This will particularly focus on specific challenges, needs, and future opportunities in contestable AI along AI value chains.

\section{Practicalities} 

We envision the workshop to be held primarily in-person, with a limited capacity for virtual participation, using an online goup chat. Plenary sessions will be streamed via videoconferencing.
To accommodate participants who may not be able to attend in person, we will allow them to present their work remotely via asynchronous video recordings 
or through live remote presentation. 
If more than six participants are online, we will group them and accommodate their participation in the group activities. Five workshop organizers will potentially be online and involved via Zoom.

To support the various activities, we will need access to a standard conference room that can accommodate up to 40 people (we envision a maximum of 30 participants in the workshop, and 10 in-person workshop organizers). Standard audiovisual equipment for the presentations and group sessions will also be required. 	

Ideally, the seating will be flexible, to cater for the presentation and group activities. Several tables to conduct the group activities will be needed. Other resources such as whiteboards, large blank papers, pens, and sticky notes can facilitate group sessions.

\section{Workshop attendance}

\subsection{Workshop participants}
We will recruit workshop participants via a call for submissions. Each author of accepted submissions will be granted a seat at the workshop. 
The call for submissions will be circulated on various HCI mailing lists and social media. Submission authors will be asked to send their submissions to the email address for the workshop before September 15th, 2024. At least two workshop organizers will review each submission. 

If we do not reach the maximum number of participants via this process, we will open participation to people who do not have any submissions. For that, we will invite interested people to send a brief statement describing one’s motivation to the workshop organizers who will then confirm the participant's spot. 
We will also directly reach out to our contacts, e.g., from civil society, and invite them to ensure participants' diversity.

We will invite submissions that can contribute to raising discussions for setting an agenda around contestable AI along the AI value chain. These submissions can focus on various sites of the AI value chain, and discuss who might want to contest what 
and towards which end-goal, when and how to contest, 
the challenges, needs, supports, and opportunities for contestation, and the limitations of current contestability works. 
The submissions should not exceed 4000 words with unlimited references and supplementary material. 

We encourage submissions from various disciplines. Not only can these submissions deal with HCI/CSCW or algorithmic work, but also policy perspectives 
and the views of advocacy organizations.

\subsection{Workshop organizers}

The diverse backgrounds of the workshop organizers allow us to cover computer science and human-computer interaction, across various continents, with expertise primarily stemming from academia but also industry and policy making.

\begin{itemize}[leftmargin=*]
    \item \textit{Agathe Balayn} is a postdoctoral researcher at Deft University of Technology. 
    Her work lies at the intersection between the technical underpinnings of AI, their operationalization in practice, and AI policy. She has conducted extensive qualitative work in organizations producing and consuming AI systems, to understand the concerns of stakeholders along the AI supply chain, the factors impacting their practices, and the harms that might arise. 
    \item \textit{Yulu Pi} is a PhD student at the Centre for Interdisciplinary Methodologies, University of Warwick. She also works on the IN-DEPTH EU AI TOOLKIT project for the Leverhulme Centre for the Future of Intelligence. 
    Her research focuses on empowering those affected by AI through explainability and contestability. Her work extends beyond technical and design issues to consider how these concepts can be incorporated into AI governance.
    \item \textit{David Gray Widder} studies how people creating AI systems think about the downstream harms their systems make possible, and the wider cultural, political, and economic logics which shape these thoughts. He is a Postdoctoral Fellow at the Digital Life Initiative at Cornell Tech. 
    He has previously conducted research at Intel Labs, Microsoft Research, and NASA’s JPL. 
    \item \textit{Kars Alfrink} is a designer, researcher, and educator working at the intersection of emerging technologies, social progress, and the built environment. He is currently a postdoctoral researcher at Delft University of Technology, focusing on contestable AI. 
    With over 15 years of experience in design, he has previously worked as a consultant, entrepreneur, and educator and has held various roles in web agencies. 
    \item \textit{Mireia Yurrita} is a PhD student at Delft University of Technology, as well as a Marie Curie fellow at the DCODE Network. Mireia’s research interests lie at the intersection of Human-AI Interaction and Algorithmic Fairness, Accountability and Transparency. Mireia has previously researched into the effect of contestability on decision subjects’ fairness perceptions, as well as, algorithmic decision subjects needs for meaningful contestability.

    \item \textit{Sohini Upadhyay} is a Computer Science PhD candidate at Harvard University. 
    She has worked on
    algorithmic recourse and XAI, and her current interests in AI contestation span microscale approaches to information needs to macroscale perspectives on socio-technical interventions and collective action. 
    She has interned at the Surveillance Technology Oversight Project 
    and the Technology for Liberty Project at the ACLU of Massachusetts. 
    \item \textit{Naveena Karusala} is a postdoctoral fellow at the Center for Research on Computation and Society at Harvard University. Her research is at the intersection of human-centered AI and the future of care work. She examines how emerging AI technologies might be designed to support more just care infrastructures that better recognize the labor, autonomy, and rights of care workers and care recipients. 
    \item \textit{Henrietta Lyons} is a PhD candidate in the School of Computing and Information Systems at the University of Melbourne. Her 
    research focuses on how contestability in algorithmic decision-making can be operationalised, and on people’s perceptions of different appeal processes for algorithmic decisions. She is interested in the societal impacts of the use of AI and the design of responsible AI systems. 
    \item \textit{Cagatay Turkay} is a Professor at the Centre for Interdisciplinary Methodologies at the University of Warwick. His research investigates the interactions between data, algorithms and people, and explores the role of interactive data visualisations and other interaction mediums such as natural language. 
    He has been 
    chairing events such as IEEE VIS, BioVis and EuroVA.
     \item \textit{Christelle Tessono} is a technology policy researcher and advocate pursuing her graduate studies at the University of Toronto. 
     She helps lead a national coalition to ban facial recognition in Canada. Her research seeks to address the relationship between digital technology and racial inequality from a policy lens. This has led her to work on projects related to political advertising on social media platforms, gig work, facial recognition technology, AI governance. 
     \item \textit{Blair Attard-Frost} is a PhD Candidate and SSHRC Joseph-Armand Bombardier Canada Graduate Scholar at the University of Toronto. Blair applies a transfeminist lens to investigate Canada's AI governance system and the contestability of AI governance practices. Drawing on 10+ years of experience working across the public sector and industry, they teach courses on AI policy and advocate for community-led AI governance. 
    \item \textit{Ujwal Gadiraju} is an Assistant Professor at 
    Delft University of Technology and a Director of the Delft “Design@Scale” AI Lab. 
    He co-leads a research line on Crowd Computing and Human-Centered AI. 
    His work lies at the intersection of HCI, AI, and information retrieval. His goal 
    is to help people far and wide by fostering meaningful reliance on AI. 
\end{itemize}

\bibliographystyle{ACM-Reference-Format}
\bibliography{main}

\end{document}